\documentclass[runningheads]{llncs}

\usepackage{main}

\usepackage{mainabbrv}

\usepackage{graphicx}
\usepackage{booktabs}
\usepackage{multirow}
\usepackage{float}
\usepackage{tikz}
\usetikzlibrary{positioning, fit, arrows.meta, calc}
\usepackage{listings}
\usepackage{adjustbox}
\usepackage{seqsplit}
\usepackage{threeparttable}

\usepackage[accsupp]{axessibility}  %

\usepackage{hyperref}

\usepackage{orcidlink}

\newif\ifanonymous
\anonymousfalse  %

\begin{document}

\title{POTATR: A Lightweight Image-to-Graph Model for Page-Level Table Extraction}

\titlerunning{POTATR for Page-Level Table Extraction}

\ifanonymous
  \author{Anonymous Authors}
  \authorrunning{Anonymous}
  \institute{Anonymous Institution}
\else
  \author{Brandon Smock \and Libin Liang \and Maxim Sokolov \and Amrit Ramesh \and Valerie Faucon-Morin \and Tayyibah Khanam \and Maury Courtland}
  \authorrunning{B.~Smock et al.}
  \institute{Kensho Technologies\\
  \email{\{brandon.smock,maury.courtland\}@kensho.com}
  }
\fi

\maketitle
\begin{abstract}
Large-scale document processing requires contextually aware table extraction (TE) that is both accurate and efficient.
Yet current approaches require billions of parameters, hundreds of autoregressive steps, or costly API inference.
Motivated by this, we introduce the Page-Object Table Transformer (POTATR), a 29M parameter image-to-graph model that extends the Table Transformer for contextualized page-level TE.
POTATR outperforms all models tested on PubTables-v2 Single Pages---including frontier MLLMs---achieving $\textrm{GriTS}_\textrm{Con}$ of 0.965 while running over 130$\times$ faster at 300$\times$ lower cost.
Further, POTATR's output is spatially grounded: every recognized element has a bounding box, enabling visual verification and geometric text assignment.
As a result, POTATR performs unified page-level TE while composing with other models, enabling extension to scanned documents via external OCR and to full-document TE via techniques like cross-page merging.
Code and models are available at \url{https://github.com/kensho-technologies/POTATR}.
\end{abstract}
  
\begin{figure}[H]
\centering
\includegraphics[width=0.92\textwidth]{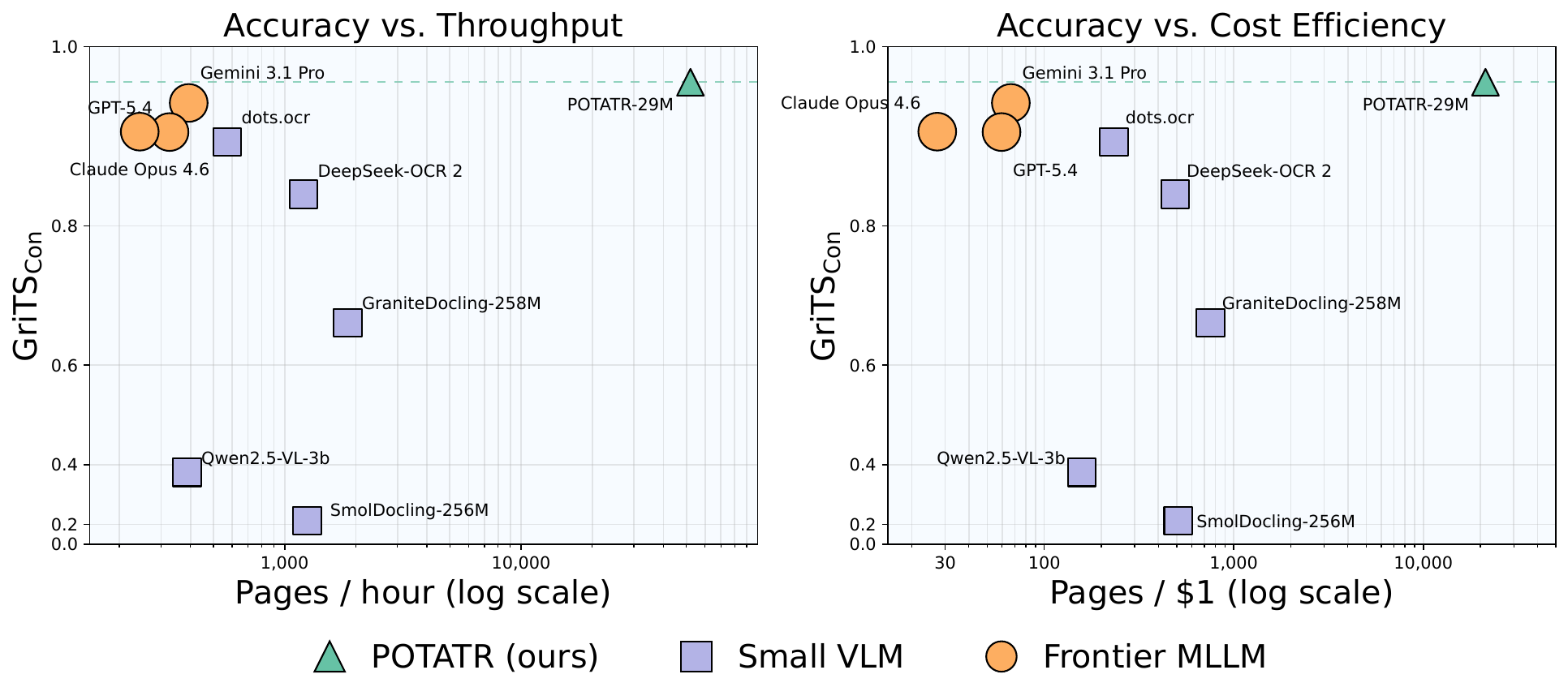}
\caption{\textbf{Accuracy vs.\ efficiency on PubTables-v2 Single Pages.} POTATR (29M parameters) achieves the highest page-level table extraction accuracy ($\textrm{GriTS}_\textrm{Con}$) while being over 130$\times$ faster and 300$\times$ cheaper than frontier MLLMs. The y-axis uses a nonlinear scale to better separate high-accuracy models.}
\label{fig:accuracy_vs_efficiency}
\end{figure}

\section{Introduction}
\label{sec:intro}

Vision-language models (VLMs) \cite{bai2025qwen2, nassar2025smoldocling, team2025granite, li2025dots} have emerged as the dominant paradigm for contextually aware table extraction (TE).
Such models frame TE as autoregressive text generation---producing HTML or markdown token-by-token.
This has the advantage of leveraging vast amounts of structured text data for training, which has been shown to lead to strong recognition performance.

However, this paradigm has several downsides.
These include inherently sequential generation, lack of structural validity guarantees, and often a lack of reliable spatial grounding for page elements.
Moreover, despite their increasing effectiveness, frontier multi-modal large language models (MLLMs) can be prohibitively costly to deploy at scale \cite{poznanski2025olmocr}. Task-specific systems, in contrast, are orders of magnitude less expensive than multi-purpose generative models \cite{luccioni2024power}.

Motivated by these limitations, we explore an alternative paradigm.
We introduce the Page-Object Table Transformer (POTATR), a lightweight 29M-parameter image-to-graph model for contextually aware page-level TE.
Rather than autoregressive text generation, POTATR frames page-level TE as parallel-decoded spatial graph prediction.
It extends TATR \cite{smock2022pubtables, smock2023aligning}---a proven architecture for table structure recognition (TSR) on cropped tables---to the page level, adding page-specific output classes and a relation head for graph prediction.
Despite being orders of magnitude smaller than VLMs, POTATR outperforms all tested models on page-level TE---including frontier MLLMs---at a fraction of the cost (\cref{fig:accuracy_vs_efficiency}).

Our contributions:\vspace{-0.3\topsep}
\begin{itemize}
    \item We introduce POTATR, a 29M-parameter model that frames page-level TE as image-to-graph prediction, jointly detecting tables, their structures, captions, and footers, with hierarchical relationships between them, and every element spatially grounded.
    \item We show that initializing from pre-trained TSR weights provides a significant advantage over training image-to-graph models from scratch, validating the choice to extend an existing TSR model for page-level TE.
    \item We show that POTATR outperforms all tested models on PubTables-v2 Single Pages---including frontier MLLMs---achieving $\textrm{GriTS}_\textrm{Con}$ of 0.965 and caption text F1 of 0.981, while running over 130$\times$ faster at 300$\times$ lower cost.
    \item Finally, we show that POTATR's modular, spatially-grounded design composes naturally with other specialized models, which enables system-level improvements on downstream tasks such as multi-page table extraction. Altogether, we believe this shows that small, non-autoregressive models remain a compelling option for deploying real-world document understanding at scale.
\end{itemize}

\section{Related Work}
\label{sec:related_work}

\subsection{Table Extraction}

Table extraction has traditionally decomposed into table detection on individual pages followed by table structure recognition (TSR) on cropped table images \cite{schreiber2017deepdesrt, prasad2020cascadetabnet, smock2022pubtables, nassar2022tableformer}.
Several works have attempted to unify these stages in a single model, including segmentation-based approaches \cite{paliwal2019tablenet, prasad2020cascadetabnet}, detection-based hierarchies \cite{zheng2021global}, and corner-edge segmentation \cite{baek2023trace}.
However, many of these methods rely on spatial heuristics or rule-based post-processing to associate detected elements with their parent tables, rather than learning these relationships directly.

The Table Transformer (TATR) \cite{smock2022pubtables, smock2023aligning} is a DETR-based \cite{carion2020end} model that achieves strong TSR on cropped tables, but is limited to processing individual tables in isolation.
POTATR extends TATR to full-page TE, leveraging its pre-trained weights while adding the capabilities needed for contextualized extraction.

Most closely related to our work, the End-to-End Table Transformer (ETT) \cite{choi2024end} extends Deformable DETR with a symmetric adjacency matrix to group detected structures by table membership.
POTATR differs in several key respects: it predicts \textit{directed} parent-child relations rather than symmetric same-table adjacency, leverages pre-trained TATR weights rather than training from scratch, includes caption and footer classes as first-class page elements, handles rotated tables directly in page context, and uses a lighter architecture (29M vs.\ $\sim$41M+ parameters).

\subsection{Image-to-Graph Models}

Image-to-graph models jointly predict objects and their relationships from images.
DocParser \cite{rausch2021docparser} applied this paradigm to document understanding, predicting hierarchical relationships between page elements.
Relationformer \cite{shit2022relationformer} proposes a general-purpose framework based on Deformable DETR \cite{zhu2020deformable}, using a dedicated relation token and relation head to predict edges between detected objects.
EGTR \cite{im2024egtr} repurposes Deformable DETR's internal query-key interactions for scene graph generation.

Page-level TE is inherently hierarchical: tables contain structures (rows, columns, headers) and are associated with captions and footers.
Unlike prior work, POTATR produces a full directed hierarchical graph for page-level TE, predicting not only parent-child edges between tables and their structures, but also between tables and their captions and footers.

\subsection{Vision-Language Models for Document Parsing}

VLMs \cite{alayrac2022flamingo, bai2025qwen2, team2023gemini} combine visual input with text prompts for multi-modal understanding.
A recent trend is toward smaller, specialized VLMs for document parsing \cite{nassar2025smoldocling, granitedocling2025, team2025granite, li2025dots, wei2025deepseek, wei2026deepseek}, which are less expensive to train and deploy.
However, even these smaller models typically contain hundreds of millions to billions of parameters and are evaluated only on isolated cropped tables.

In practice, document parsing often involves modular pipelines combining multiple specialized models (e.g., Docling \cite{nassar2025smoldocling}).
Our work demonstrates that a much smaller specialized model (29M parameters) can outperform VLMs on page-level TE when paired with accurate text extraction.
Unlike VLMs that entangle structure detection with text generation, POTATR's modular design separates these concerns---yielding a component that can be integrated into broader document parsing pipelines while remaining independently upgradeable.

\section{Method}
\label{sec:method}

\begin{figure}[tb]
\centering
\resizebox{\textwidth}{!}{%
\begin{tikzpicture}[
    node distance=0.5cm and 0.7cm,
    box/.style={rectangle, draw, rounded corners=5pt, minimum height=1.1cm, minimum width=2.7cm, align=center, font=\large, inner sep=5pt},
    myshadow/.style={preaction={fill=gray!40, transform canvas={xshift=2pt, yshift=-2pt}}},
    pretrained/.style={box, fill=gray!4, draw=gray!90, line width=1.2pt, myshadow},
    newmodule/.style={box, fill=teal!6, draw=teal!60, line width=1.2pt, myshadow},
    expanded/.style={box, fill=orange!6, draw=orange!60, line width=1.2pt, dash pattern=on 5pt off 3pt, myshadow},
    output/.style={box, fill=white, draw=none, minimum width=1.8cm},
    inputbox/.style={box, fill=gray!3, draw=gray!40},
    mainarrow/.style={->, >=stealth, line width=1.6pt, color=gray!70},
    brancharrow/.style={->, >=stealth, line width=0.9pt, color=gray!65},
    groupbox/.style={draw=gray!35, dashed, rounded corners=6pt, inner sep=10pt},
]

\node[inner sep=1pt, draw=gray!60, line width=0.4pt, myshadow] (input) {\includegraphics[width=1.92cm, height=2.5cm]{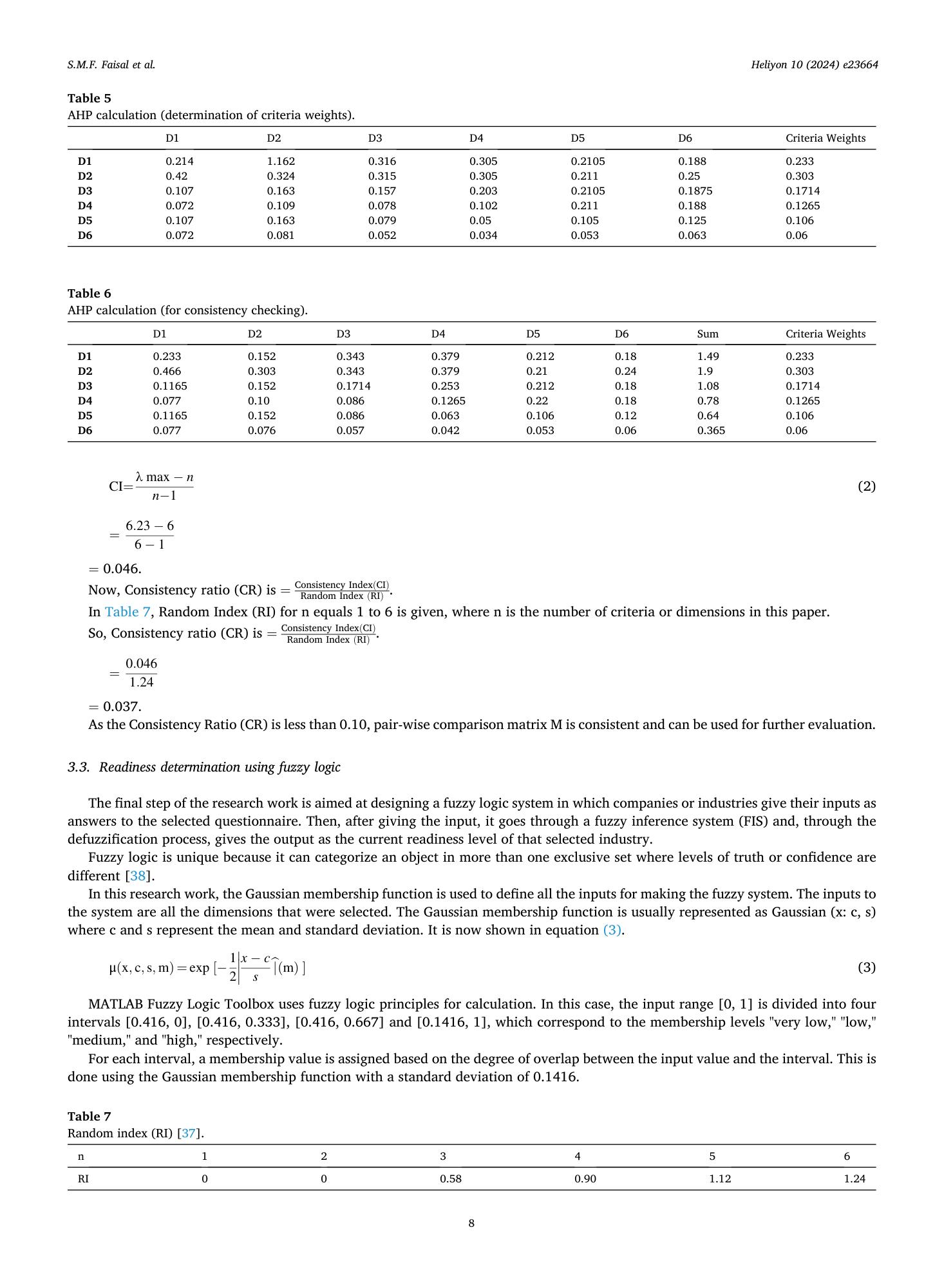}};
\node[font=\small, below=0.1cm of input] (inputlabel) {Page Image};

\node[pretrained, right=1.0cm of input, inner sep=8pt, rounded corners=8pt, font=\large\bfseries] (backbone) {ResNet-18\\Backbone};

\node[pretrained, right=0.6cm of backbone, inner sep=8pt, rounded corners=8pt, font=\large\bfseries] (encoder) {Transformer\\Encoder};

\node[pretrained, right=0.6cm of encoder, inner sep=8pt, rounded corners=8pt, font=\large\bfseries] (decoder) {Transformer\\Decoder};

\node[draw=gray!40, dashed, rounded corners=8pt, line width=0.5pt, inner sep=8pt, fit=(backbone)(encoder)(decoder), label={[font=\small, text=gray!80]below:Base Model}] (basegroup) {};

\node[box, fill=orange!6, draw=orange!60, line width=1.2pt, dash pattern=on 5pt off 3pt, above=1.0cm of decoder, minimum width=2.4cm, minimum height=0.6cm, font=\small] (queries) {Object Queries};

\node[expanded, above right=0.7cm and 1.1cm of decoder, minimum height=0.6cm, minimum width=2.4cm, font=\normalsize] (classhead) {Class Head};
\node[pretrained, right=1.1cm of decoder, minimum height=0.6cm, minimum width=2.4cm, font=\normalsize] (bboxhead) {BBox Head};
\node[newmodule, below right=0.7cm and 1.1cm of decoder, minimum height=0.6cm, minimum width=2.4cm, font=\normalsize] (relhead) {Relation Head};

\node[output, right=0.5cm of classhead, font=\small, minimum width=2.8cm] (classes) {table, row, column,\\caption, footer, \ldots};
\node[output, right=0.5cm of bboxhead, font=\small] (bboxes) {$(x, y, w, h)$};

\node[output, right=0.5cm of relhead, minimum width=2.8cm, minimum height=1.2cm] (relations) {};
\node[circle, fill=gray!30, draw=gray!50, inner sep=2pt, font=\tiny] at ([xshift=-0.9cm, yshift=0.2cm]relations.center) (g1) {T};
\node[circle, fill=white, draw=gray!40, inner sep=1.5pt, font=\tiny] at ([xshift=-0.1cm, yshift=0.3cm]relations.center) (g2) {R};
\node[circle, fill=white, draw=gray!40, inner sep=1.5pt, font=\tiny] at ([xshift=0.3cm, yshift=-0.1cm]relations.center) (g3) {R};
\node[circle, fill=white, draw=gray!40, inner sep=1.5pt, font=\tiny] at ([xshift=-0.5cm, yshift=-0.3cm]relations.center) (g4) {C};
\draw[->, >=stealth, thin, gray!60] (g1) -- (g2);
\draw[->, >=stealth, thin, gray!60] (g1) -- (g3);
\draw[->, >=stealth, thin, gray!60] (g1) -- (g4);

\draw[mainarrow] (input) -- (backbone);
\draw[mainarrow] (backbone) -- (encoder);
\draw[mainarrow] (encoder) -- (decoder);

\draw[brancharrow] (queries.south) -- (decoder.north);

\draw[brancharrow] (decoder.east) -- ++(0.7,0) |- (classhead.west);
\draw[brancharrow] (decoder.east) -- (bboxhead.west);
\draw[brancharrow] (decoder.east) -- ++(0.7,0) |- (relhead.west);

\draw[brancharrow] (classhead) -- (classes);
\draw[brancharrow] (bboxhead) -- (bboxes);
\draw[brancharrow] (relhead) -- (relations);

\coordinate (legend_center) at ($(current bounding box.south west)!0.5!(current bounding box.south east) + (0.7,-0.6)$);
\node[rectangle, draw=gray!90, fill=gray!4, line width=0.6pt, rounded corners=3pt, minimum height=0.35cm, font=\normalsize, inner sep=5pt] at ([xshift=-1.8cm]legend_center) (leg1) {\vphantom{p}Same};
\node[rectangle, draw=orange!60, fill=orange!6, line width=0.6pt, dash pattern=on 5pt off 3pt, rounded corners=3pt, minimum height=0.35cm, font=\normalsize, inner sep=5pt] at (legend_center) (leg2) {Expanded};
\node[rectangle, draw=teal!60, fill=teal!6, line width=0.6pt, rounded corners=3pt, minimum height=0.35cm, font=\normalsize, inner sep=5pt] at ([xshift=1.8cm]legend_center) (leg3) {\vphantom{p}New};

\node[font=\small, text=gray!80] at ([xshift=-3.2cm]legend_center) (legendlabel) {Legend:};
\node[draw=gray!40, dashed, rounded corners=6pt, line width=0.5pt, inner sep=6pt, fit=(legendlabel)(leg1)(leg2)(leg3)] (legendbox) {};

\end{tikzpicture}%
}
\caption{\textbf{POTATR architecture.} POTATR extends TATR with 125 additional object queries, 10 new page-level object classes, and a relation head that predicts parent-child relationships between detected objects. \textit{Same} (gray): modules unchanged from pre-trained TATR. \textit{Expanded} (orange, dashed): modules with pre-trained components plus additional randomly initialized parts. \textit{New} (teal): modules added in POTATR and randomly initialized. The relation head takes pairs of decoder embeddings and predicts directed edges, producing a hierarchical graph over detected page objects.}
\label{fig:architecture_v2}
\end{figure}
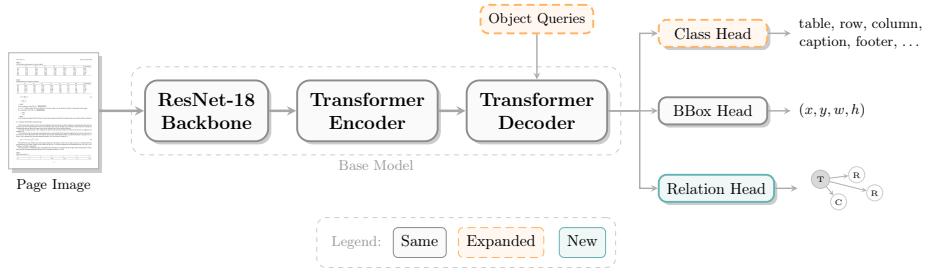

\subsection{POTATR Architecture}
\label{sec:potatr}

We extend the Table Transformer (TATR) \cite{smock2023aligning} from a model for table structure recognition on individual cropped tables to a model for full-page table extraction, which we call the Page-Object Table Transformer (POTATR).

The original TATR model predicts 6 object classes: \emph{table}, \emph{table column}, \emph{table row}, \emph{table column header}, \emph{table projected row header}, and \emph{table spanning cell}.
POTATR adds 2 additional page-level classes: \emph{table caption} and \emph{table footer}.
Because a table can be rotated within a page, POTATR also adds 8 additional classes that are the rotated counterparts of the first 8 (\ie \emph{table rotated} and \emph{table column rotated}), for a total of 16 classes.
Since page images can have more objects to detect, we also double the number of object queries from 125 to 250.

POTATR adds a relation head similar to Relationformer \cite{shit2022relationformer}, making it an image-to-graph model.
We frame relation prediction as a binary classification problem that predicts whether object 1 is the parent of object 2.
We define the table object to always be the parent, while its structures (rows, columns, etc.), caption, and footer objects are children.
This serves the purpose of associating captions and footers with their table, as well as grouping table structures that belong to the same table.
The grouping of table structures using explicit relation prediction departs from the approach used by TATR, which instead uses bounding box overlap with a parent table object to determine the grouping, similar to DocParser \cite{rausch2021docparser}.

POTATR as an image-to-graph architecture differs from Relationformer in two key ways.
First, POTATR is based on the original DETR architecture \cite{carion2020end}, same as TATR.
Relationformer and other image-to-graph models adopt Deformable DETR \cite{zhu2020deformable} as their base architecture.
Second, POTATR simplifies the relation head, dropping the use of a relation token.
We hypothesize that the relation token is unnecessary when using full self-attention as in the original DETR architecture.
The relation head is implemented as a three-layer MLP. The total parameter count for POTATR is 29M.

The main purpose for extending TATR into POTATR rather than adopting an existing image-to-graph model such as Relationformer or EGTR \cite{im2024egtr} is to leverage TATR's pre-trained weights.
We hypothesized that this would significantly boost model performance and confirm this with a small-scale experiment in \cref{sec:graph_model_comparison}.

\subsection{Training}
\label{sec:potatr_training}

We train and evaluate on PubTables-v2 \cite{smock2025pubtablesv2}, a large-scale dataset for page- and document-level TE sourced from over one million PubMed articles.
PubTables-v2 (PTv2) contains multiple collections, including two collections with full pages as input: PTv2 Single Pages (468k pages with 548k tables annotated with hierarchical relationships between tables, captions, and footers), and PTv2 Full Documents (9k documents containing 137k pages and $\sim$25k tables, including 9.5k multi-page tables, annotated with HTML for TSR and bounding boxes for table detection).

We train POTATR in two stages.
In the first stage, we train for 100 epochs on the Single Pages collection without any change to the data distribution.
In the second stage, we train for an additional 35 epochs where each epoch contains an additional 2,000 table-free document pages from the Full Documents collection, as well as repeats the 10k largest tables in the Single Pages collection with different augmentation.
These additions are intended to address the long tail of the data distribution.

Training is conducted on 8 Nvidia T4 GPUs with a batch size of 2 per GPU, for an effective batch size of 16.
The initial learning rate is 0.00005 (5e-5) with a learning rate gamma of 0.9 applied every 2 epochs.
We use the default hyperparameters from TATR \cite{smock2022pubtables}, except the weight on the ``no object'' class, \texttt{eos\_coef}, which we set to 0.1.
There is also an additional relation loss with a weight of 0.05.

Most weights are initialized from \texttt{TATR-v1.1-Pub} \cite{smock2023aligning}, a pre-trained model for TSR on cropped tables.
The additional relation head, expanded class head, and the 125 additional object queries are randomly initialized.

\section{Experiments}
\label{sec:experiments}

\subsection{Metrics}
\label{sec:metrics}

For page-level table extraction, we evaluate using the page-level generalizations of GriTS \cite{smock2023grits} and TEDS \cite{zhong2020image} for TE, proposed by Smock \etal \cite{smock2025pubtablesv2}. 
At the page-level for TE, it is no longer assumed that there is exactly one predicted table evaluated against one ground truth table, as in the traditional TSR task.
Instead, each metric compares two \emph{sets} of tables (predicted and ground truth) and uses the Hungarian algorithm to determine the one-to-one correspondence that maximizes their aggregate F1 score, with GriTS or TEDS used to determine the score for all candidate matches between tables.
This flexibility enables evaluation for methods that do not produce a bounding box prediction for a table's location.
In our case, we apply the Hungarian matching within a single page, although the same approach can be applied across an entire document.

For object detection we use the standard COCO metrics and for evaluating graph prediction, we use the edge F1 metric from DocParser \cite{rausch2021docparser}.
We use an IoU threshold of 0.8 to determine if the predicted nodes for each edge are a true positive.
For caption text accuracy, we use character-level F1 based on the longest common subsequence (LCS) between predicted and ground truth caption strings.

\subsection{Image-to-Graph Model Comparison}
\label{sec:graph_model_comparison}

\begin{table}[tb]
\caption{\textbf{Image-to-graph model comparison.} To validate the choice to base POTATR on the pre-trained TATR model, we train each candidate image-to-graph model for 10 epochs on the PubTables-v2 Single Pages collection.
Interestingly, the benefits of leveraging pre-trained weights from \texttt{TATR-v1.1-Pub} \cite{smock2023aligning} extend not only to classifying the new additional object classes but also to the novel relation prediction task (edge F1), which is not part of the original TATR model.}
\centering
\begin{tabular}{l r r r r r r}
\toprule
\textbf{Model} & $\textbf{AP50}$ & $\textbf{AP75}$ & $\textbf{AP}$ & $\textbf{GriTS}_{\textbf{Top}}$ & $\textbf{GriTS}_{\textbf{Con}}$ & $\textbf{Edge F1}$ \\
\midrule
Relationformer \cite{shit2022relationformer} & 0.808 & 0.523 & 0.484 & 0.852 & 0.840 & 0.339 \\
EGTR \cite{im2024egtr} & 0.791 & 0.599 & 0.549 & 0.850 & 0.837 & 0.707 \\
\textbf{POTATR (ours)} & \textbf{0.904} & \textbf{0.755} & \textbf{0.698} & \textbf{0.902} & \textbf{0.892} & \textbf{0.746} \\
\bottomrule
\end{tabular}
\label{tab:graph_model_comparison}
\end{table}

To validate which modeling approach to adopt for our full-scale training experiments, we compare POTATR against Relationformer \cite{shit2022relationformer} and EGTR \cite{im2024egtr} in a small-scale experiment.
We match the parameters of Relationformer and EGTR to those of POTATR and use each model's default losses and loss weights.
All models are trained on 8 Nvidia T4 GPUs with a batch size of 2 per GPU (effective batch size 16) for 10 epochs on the PubTables-v2 Single Pages collection, with an initial learning rate of 0.00005 and a learning rate gamma of 0.66.

POTATR outperforms both alternatives across all metrics (\cref{tab:graph_model_comparison}).
Interestingly, POTATR outperforms Relationformer and EGTR not only on object detection but also on relation prediction, despite the relation head being an entirely new component.
This validates the choice to extend pre-trained TATR with a relation head rather than adopting an existing image-to-graph architecture to train from scratch.

\subsection{Page-Level Table Extraction}\label{sec:page_level_te}

\begin{table}[tb]
\caption{\textbf{Page-level table extraction.} Evaluation results on the PubTables-v2 Single Pages collection. POTATR is a 29M-parameter model; all other models are evaluated zero-shot. The best scores among VLMs/MLLMs are underlined; the best overall scores are bolded.}
\centering
\begin{adjustbox}{width=\textwidth}
\begin{tabular}{l r r r r r r}
\toprule
\textbf{Model} & $\textbf{GriTS}_{\textbf{Top}}$ & $\textbf{GriTS}_{\textbf{Con}}$ & $\textbf{Acc}_{\textbf{Top}}$ & $\textbf{Acc}_{\textbf{Con}}$ & $\textbf{TEDS}_{\textbf{S}}$ & $\textbf{TEDS}$ \\
\midrule
SmolDocling-256M \cite{nassar2025smoldocling} & 0.2502 & 0.2177 & 0.2800 & 0.0298 & 0.5980 & 0.5250 \\
GraniteDocling-258M \cite{granitedocling2025} & 0.7252 & 0.6670 & 0.3794 & 0.0829 & 0.6685 & 0.6155 \\
Qwen2.5-VL-3b \cite{bai2025qwen2} & 0.4672 & 0.3803 & 0.3651 & 0.0619 & 0.2859 & 0.2632 \\
Granite-Vision-3.2-2b \cite{team2025granite} & 0.8015 & 0.7480 & 0.4245 & 0.0331 & 0.8665 & 0.8092 \\
DeepSeek-OCR \cite{wei2025deepseek} & 0.8629 & 0.8207 & 0.6184 & 0.1671 & 0.7996 & 0.7639 \\
DeepSeek-OCR 2 \cite{wei2026deepseek} & 0.8839 & 0.8386 & 0.5258 & 0.1120 & 0.7688 & 0.7263 \\
dots.ocr \cite{li2025dots} & 0.9241 & 0.8991 & \underline{0.6425} & 0.1872 & 0.9168 & 0.8743 \\
\midrule
Claude Opus 4.6 & 0.9146 & 0.9106 & 0.4934 & 0.2713 & 0.9073 & 0.8945 \\
GPT-5.4 & 0.9208 & 0.9102 & 0.5461 & 0.2074 & 0.9084 & 0.8874 \\
Gemini 3.1 Pro & \underline{0.9500} & \underline{0.9418} & 0.6348 & \underline{0.3524} & \underline{0.9392} & \underline{0.9250} \\
\midrule
POTATR + EasyOCR \cite{jaidedai2024easyocr} & 0.9555 & 0.7831 & 0.6632 & 0.0019 & 0.9478 & 0.8175 \\
POTATR + PaddleOCR \cite{cui2025paddleocr} & 0.9614 & 0.8768 & 0.6953 & 0.0060 & 0.9545 & 0.8780 \\
POTATR + docTR \cite{doctr2021} & 0.9606 & 0.8809 & 0.6724 & 0.0684 & 0.9511 & 0.8971 \\
POTATR + DT & \textbf{0.9674} & \textbf{0.9647} & \textbf{0.7012} & \textbf{0.6447} & \textbf{0.9576} & \textbf{0.9516} \\
\bottomrule
\end{tabular}
\end{adjustbox}
\label{tab:page_level_TSR_results}
\end{table}

In the next experiment, we evaluate the fully-trained version of POTATR on page-level table extraction.
We compare against seven specialized document VLMs and three frontier MLLMs, all evaluated zero-shot.

Unlike VLMs, POTATR separates structure detection from text recognition: its structure predictions are combined with text either extracted directly from the document (DT), as would be present in digitally-native documents, or from a separate OCR model, as needed for scanned documents.
In addition to text extracted directly from the document, we evaluate POTATR with text extracted by three lightweight OCR models (EasyOCR \cite{jaidedai2024easyocr}, PaddleOCR \cite{cui2025paddleocr}, and docTR \cite{doctr2021}).

Results are in \cref{tab:page_level_TSR_results}.
POTATR with direct text achieves $\textrm{GriTS}_\textrm{Con}$ of 0.965---outperforming even the strongest frontier MLLM despite being a 29M-parameter model.
The best frontier MLLM (Gemini 3.1 Pro, $\textrm{GriTS}_\textrm{Con}$ 0.942) outperforms POTATR paired with traditional OCR.
The OCR progression (0.783 $\rightarrow$ 0.877 $\rightarrow$ 0.881 $\rightarrow$ 0.965) demonstrates that POTATR's structure recognition is already highly accurate; remaining errors are dominated by text extraction quality and are independently fixable.

\subsection{Caption Text Accuracy}
\label{sec:caption_accuracy}

\begin{table}[tb]
\caption{\textbf{Caption text content accuracy.} Character-level LCS-based F1 on PubTables-v2 Single Pages. POTATR detects caption bounding boxes and assigns text via direct extraction (DT) or OCR; MLLMs are prompted to output caption text alongside table structure.}
\centering
\begin{tabular}{l r r r}
\toprule
\textbf{Model} & \textbf{F1} & \textbf{Precision} & \textbf{Recall} \\
\midrule
Claude Opus 4.6 & 0.9611 & 0.9866 & 0.9370 \\
GPT-5.4 & 0.9658 & 0.9743 & 0.9575 \\
Gemini 3.1 Pro & 0.9774 & \textbf{0.9882} & 0.9669 \\
\midrule
POTATR + EasyOCR & 0.8906 & 0.8547 & 0.9297 \\
POTATR + PaddleOCR & 0.8959 & 0.8769 & 0.9156 \\
POTATR + docTR & 0.9576 & 0.9685 & 0.9469 \\
POTATR + DT & \textbf{0.9807} & 0.9824 & \textbf{0.9790} \\
\bottomrule
\end{tabular}
\label{tab:caption_accuracy}
\end{table}

Caption extraction is a natural byproduct of POTATR's spatial detection: detected caption bounding boxes are populated with text from the same source used for table cells.
We evaluate caption text accuracy using character-level LCS-based F1 (\cref{tab:caption_accuracy}).
POTATR with direct text achieves F1 of 0.981 with balanced precision and recall, slightly exceeding the best frontier MLLM (Gemini 3.1 Pro, F1 0.977).
MLLMs tend toward higher precision but lower recall---they are accurate when they output captions but occasionally omit them.
POTATR's detection-based approach ensures systematic caption coverage without requiring explicit prompting.

\subsection{Inference Cost and Throughput}
\label{sec:cost}

\begin{table}[tb]
\caption{\textbf{Inference cost and throughput.} Measured on the PubTables-v2 Single Pages test set. All local models use a single NVIDIA A10G GPU (24\,GB); VLMs use batched inference. Frontier MLLMs are priced via API. Local cost is based on an on-demand cloud instance priced at \$2.45/hr.}
\centering
\begin{tabular}{l r r r}
\toprule
\textbf{Model} & \textbf{Params} & \textbf{Pages/hr} & \textbf{\$/1M pages} \\
\midrule
Claude Opus 4.6 & — & 244 & \$36,680 \\
GPT-5.4 & — & 326 & \$16,800 \\
Gemini 3.1 Pro & — & 393 & \$15,000 \\
Qwen2.5-VL-3b & 3B & 386 & \$6,336 \\
dots.ocr \cite{li2025dots} & 3B & 572 & \$4,281 \\
DeepSeek-OCR 2 \cite{wei2026deepseek} & 3B & 1,201 & \$2,038 \\
SmolDocling-256M & 256M & 1,247 & \$1,963 \\
GraniteDocling-258M & 258M & 1,847 & \$1,325 \\
POTATR & 29M & \textbf{52,031} & \textbf{\$47} \\
\bottomrule
\end{tabular}
\label{tab:cost}
\end{table}

We compare inference cost and throughput for page-level TE (\cref{tab:cost}).
POTATR processes over 52k pages per hour on a single A10G GPU---over 130$\times$ the throughput of the fastest frontier MLLM (Gemini 3.1 Pro at 393 pages/hour) and 28$\times$ faster than the fastest local VLM (GraniteDocling at 1,847 pages/hour).
At under \$50 per million pages, POTATR is roughly 300$\times$ cheaper than the most economical frontier MLLM and 28$\times$ cheaper than the most economical local VLM.
These differences make POTATR practical for large-scale batch processing where even locally-deployed VLMs would be costly.

\begin{figure}[tb]
\centering
\includegraphics[width=\textwidth]{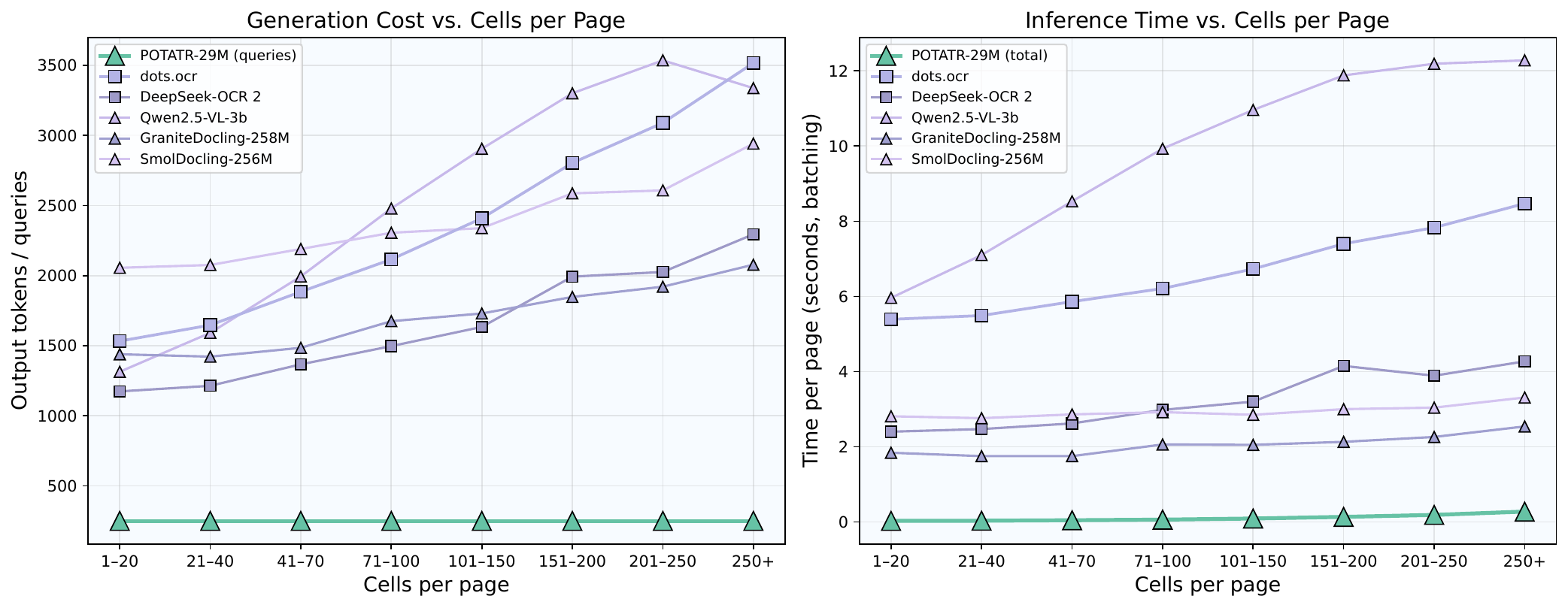}
\caption{\textbf{Generation cost vs.\ page table cell volume.} Left: output tokens (or fixed queries for POTATR) as a function of cells per page. Right: inference time per page with batching. POTATR uses a constant 250 queries regardless of page content; autoregressive VLMs generate tokens proportional to table size, leading to linearly growing cost and latency.}
\label{fig:generation_cost}
\end{figure}

This efficiency gap widens as a page's table cell count increases (\cref{fig:generation_cost}).
POTATR's inference time is constant at ${\sim}21$\,ms regardless of cells per page. Only post-processing (text assignment)---a relatively trivial operation---scales with page table cell count.
In contrast, autoregressive VLMs generate output tokens proportional to table size, with inference time growing from 1.8\,s to 12\,s across the same range.
This makes POTATR's throughput advantage even more pronounced on table-heavy document types, even compared with the most economical VLMs running in batch mode.

\section{Discussion}
\label{sec:discussion}

\subsection{Modularity as a Feature}

POTATR's modular design separates structure detection from text recognition, making errors decomposable and each component independently improvable.

POTATR with direct text achieves $\textrm{GriTS}_\textrm{Con}$ of 0.965, leaving a total error of 0.035.
The gap from POTATR + docTR (0.881) to POTATR + DT (0.965) is 0.084---more than twice the remaining error with perfect text.
This indicates that text recognition error dominates structure recognition error in the POTATR + docTR result, and that structure recognition is near its ceiling.
The OCR progression (EasyOCR 0.783 $\rightarrow$ PaddleOCR 0.877 $\rightarrow$ docTR 0.881 $\rightarrow$ DT 0.965) confirms that a system for page-level table extraction with POTATR automatically benefits from better text extraction without retraining.

VLMs entangle structure and text in a single autoregressive generation process.
When a VLM produces incorrect output, it can be difficult to fully attribute the underlying cause of the error.
It is also challenging to remedy each issue in isolation, as improving structure recognition risks harming text recognition and vice versa.
POTATR allows each component to be improved, replaced, or upgraded independently.

\subsection{From Page-Level to Document-Level TE}\label{sec:document-te}

In practice, documents---not individual pages---are the unit of processing.
Assessing page-level models in a document-level context is therefore important for understanding real-world applicability.

PubTables-v2 includes a Full Documents benchmark.
But this benchmark is not intended to assess TE performance across a typical distribution of documents.
Instead, it is designed specifically to augment the Single Pages benchmark by focusing on documents that contain multi-page tables, which is the case where page-level models are most likely to fail.

Since POTATR is trained on single pages, it has never seen tables that are split across page boundaries.
Optimizing POTATR for full documents with multi-page tables is beyond the scope of this work.
However, we do think it is of interest to assess the gap between fully single-page TE and complicated document-level TE involving multi-page tables.

We evaluate POTATR in this context in the Appendix (\cref{sec:merging_results}).
Within a system designed to incorporate a page-level model, POTATR performs well even on multi-page tables, closing 63--77\% of the gap to frontier MLLMs at roughly two orders of magnitude lower inference cost (\cref{tab:cost}).
Further optimization of POTATR for out-of-distribution multi-page content, and extending the pipeline to handle within-page table part merging, could narrow the remaining gap.

\section{Conclusion}
\label{sec:conclusion}

We introduced POTATR, a 29M-parameter image-to-graph model extending the Table Transformer for page-level table extraction, which aims to address some of the shortcomings and prohibitive costs associated with autoregressive VLMs.
POTATR outperforms all tested models on the PubTables-v2 Single Pages benchmark, including frontier MLLMs, achieving a $\textrm{GriTS}_\textrm{Con}$ of 0.965 while running over 130$\times$ faster at 300$\times$ lower cost.
POTATR also achieves similar performance to frontier MLLMs on caption text recognition, with a text F1 score of 0.981.
Notably, its modular design, paired with its spatially-grounded recognition output, composes well with other small specialized models.
Overall, these results suggest small, non-autoregressive models remain a compelling option for real-world page-level table extraction, where their constant inference time regardless of table size offers a fundamental scaling advantage.

\section{Limitations}
\label{sec:limitations}

POTATR is trained and evaluated on a dataset of English scientific articles from PubMed; generalization to other document types and languages remains untested.
However, TATR---the model that POTATR extends---has been demonstrated to generalize across a variety of document types \cite{smock2023aligning}, suggesting that domain adaptation is feasible.

\bibliographystyle{splncs04}
\bibliography{main}

\begin{thebibliography}{10}
\providecommand{\url}[1]{\texttt{#1}}
\providecommand{\urlprefix}{URL }
\providecommand{\doi}[1]{https://doi.org/#1}

\bibitem{alayrac2022flamingo}
Alayrac, J.B., Donahue, J., Luc, P., Miech, A., Barr, I., Hasson, Y., Lenc, K.,
  Mensch, A., Millican, K., Reynolds, M., et~al.: Flamingo: a visual language
  model for few-shot learning. Advances in neural information processing
  systems  \textbf{35},  23716--23736 (2022)

\bibitem{baek2023trace}
Baek, Y., Nam, D., Surh, J., Shin, S., Kim, S.: {TRACE}: table reconstruction
  aligned to corner and edges. In: International Conference on Document
  Analysis and Recognition. pp. 472--489. Springer (2023)

\bibitem{bai2025qwen2}
Bai, S., Chen, K., Liu, X., Wang, J., Ge, W., Song, S., Dang, K., Wang, P.,
  Wang, S., Tang, J., et~al.: Qwen2.5-{VL} technical report. arXiv preprint
  arXiv:2502.13923  (2025)

\bibitem{carion2020end}
Carion, N., Massa, F., Synnaeve, G., Usunier, N., Kirillov, A., Zagoruyko, S.:
  End-to-end object detection with transformers. In: ECCV. pp. 213--229 (2020)

\bibitem{choi2024end}
Choi, Y.Y., Kim, T., Kim, N., Lee, T., Joe, S.: End to end table transformer.
  In: International Conference on Document Analysis and Recognition. pp.
  331--345. Springer (2024)

\bibitem{cui2025paddleocr}
Cui, C., Sun, T., Lin, M., Gao, T., Zhang, Y., Liu, J., Wang, X., Zhang, Z.,
  Zhou, C., Liu, H., et~al.: Paddle{OCR} 3.0 technical report. arXiv preprint
  arXiv:2507.05595  (2025)

\bibitem{granitedocling2025}
IBM-Granite: Granite {D}ocling.
  \url{https://huggingface.co/ibm-granite/granite-docling-258M} (2025)

\bibitem{im2024egtr}
Im, J., Nam, J., Park, N., Lee, H., Park, S.: E{GTR}: Extracting graph from
  transformer for scene graph generation. In: Proceedings of the IEEE/CVF
  Conference on Computer Vision and Pattern Recognition. pp. 24229--24238
  (2024)

\bibitem{jaidedai2024easyocr}
Jaided{AI}: Easy{OCR} (09 2024), \url{https://github.com/JaidedAI/EasyOCR}

\bibitem{li2025dots}
Li, Y., Yang, G., Liu, H., Wang, B., Zhang, C.: dots.ocr: {M}ultilingual
  document layout parsing in a single vision-language model. arXiv preprint
  arXiv:2512.02498  (2025)

\bibitem{luccioni2024power}
Luccioni, A.S., Jernite, Y., Strubell, E.: Power hungry processing: Watts
  driving the cost of ai deployment? In: Proceedings of the 2024 ACM Conference
  on Fairness, Accountability, and Transparency (FAccT). pp. 85--99. ACM (2024)

\bibitem{doctr2021}
Mindee: doc{TR}: Document text recognition.
  \url{https://github.com/mindee/doctr} (2021)

\bibitem{nassar2022tableformer}
Nassar, A., Livathinos, N., Lysak, M., Staar, P.: Tableformer: Table structure
  understanding with transformers. In: Proceedings of the IEEE/CVF Conference
  on Computer Vision and Pattern Recognition. pp. 4614--4623 (2022)

\bibitem{nassar2025smoldocling}
Nassar, A., Marafioti, A., Omenetti, M., Lysak, M., Livathinos, N., Auer, C.,
  Morin, L., de~Lima, R.T., Kim, Y., Gurbuz, A.S., et~al.: Smol{D}ocling: An
  ultra-compact vision-language model for end-to-end multi-modal document
  conversion. arXiv preprint arXiv:2503.11576  (2025)

\bibitem{paliwal2019tablenet}
Paliwal, S.S., Vishwanath, D., Rahul, R., Sharma, M., Vig, L.: Tablenet: Deep
  learning model for end-to-end table detection and tabular data extraction
  from scanned document images. In: 2019 international conference on document
  analysis and recognition (ICDAR). pp. 128--133. IEEE (2019)

\bibitem{poznanski2025olmocr}
Poznanski, J., Rangapur, A., Borchardt, J., Dunkelberger, J., Huff, R., Lin,
  D., Wilhelm, C., Lo, K., Soldaini, L.: olmocr: Unlocking trillions of tokens
  in pdfs with vision language models. arXiv preprint arXiv:2502.18443  (2025)

\bibitem{prasad2020cascadetabnet}
Prasad, D., Gadpal, A., Kapadni, K., Visave, M., Sultanpure, K.:
  Cascade{T}ab{N}et: An approach for end to end table detection and structure
  recognition from image-based documents. In: Proceedings of the IEEE/CVF
  conference on computer vision and pattern recognition workshops. pp. 572--573
  (2020)

\bibitem{rausch2021docparser}
Rausch, J., Martinez, O., Bissig, F., Zhang, C., Feuerriegel, S.: Docparser:
  Hierarchical document structure parsing from renderings. In: Proceedings of
  the AAAI Conference on Artificial Intelligence. vol.~35, pp. 4328--4338
  (2021)

\bibitem{schreiber2017deepdesrt}
Schreiber, S., Agne, S., Wolf, I., Dengel, A., Ahmed, S.: Deep{D}e{STR}: Deep
  learning for detection and structure recognition of tables in document
  images. In: 2017 14th IAPR international conference on document analysis and
  recognition (ICDAR). vol.~1, pp. 1162--1167. IEEE (2017)

\bibitem{shit2022relationformer}
Shit, S., Koner, R., Wittmann, B., Paetzold, J., Ezhov, I., Li, H., Pan, J.,
  Sharifzadeh, S., Kaissis, G., Tresp, V., et~al.: Relationformer: A unified
  framework for image-to-graph generation. In: European Conference on Computer
  Vision. pp. 422--439. Springer (2022)

\bibitem{smock2025pubtablesv2}
Smock, B., Faucon-Morin, V., Sokolov, M., Liang, L., Khanam, T., Ramesh, A.,
  Courtland, M.: Pub{T}ables-v2: A new large-scale dataset for full-page and
  multi-page table extraction. arXiv preprint arXiv:2512.10888  (2025)

\bibitem{smock2022pubtables}
Smock, B., Pesala, R., Abraham, R.: Pub{T}ables-1{M}: Towards comprehensive
  table extraction from unstructured documents. In: Proceedings of the IEEE/CVF
  Conference on Computer Vision and Pattern Recognition (CVPR). pp. 4634--4642
  (June 2022)

\bibitem{smock2023aligning}
Smock, B., Pesala, R., Abraham, R.: Aligning benchmark datasets for table
  structure recognition. In: International Conference on Document Analysis and
  Recognition. pp. 371--386. Springer (2023)

\bibitem{smock2023grits}
Smock, B., Pesala, R., Abraham, R.: Gri{TS}: Grid table similarity metric for
  table structure recognition. In: International Conference on Document
  Analysis and Recognition. pp. 535--549. Springer (2023)

\bibitem{team2023gemini}
Team, G., Anil, R., Borgeaud, S., Alayrac, J.B., Yu, J., Soricut, R.,
  Schalkwyk, J., Dai, A.M., Hauth, A., Millican, K., et~al.: Gemini: a family
  of highly capable multimodal models. arXiv preprint arXiv:2312.11805  (2023)

\bibitem{team2025granite}
Team, G.V., Karlinsky, L., Arbelle, A., Daniels, A., Nassar, A., Alfassi, A.,
  Wu, B., Schwartz, E., Joshi, D., Kondic, J., et~al.: Granite vision: a
  lightweight, open-source multimodal model for enterprise intelligence. arXiv
  preprint arXiv:2502.09927  (2025)

\bibitem{wei2025deepseek}
Wei, H., Sun, Y., Li, Y.: Deepseek-ocr: Contexts optical compression. arXiv
  preprint arXiv:2510.18234  (2025)

\bibitem{wei2026deepseek}
Wei, H., Sun, Y., Li, Y.: Deepseek-ocr 2: Visual causal flow. arXiv preprint
  arXiv:2601.20552  (2026)

\bibitem{zheng2021global}
Zheng, X., Burdick, D., Popa, L., Zhong, X., Wang, N.X.R.: Global table
  extractor (gte): A framework for joint table identification and cell
  structure recognition using visual context. In: Proceedings of the IEEE/CVF
  winter conference on applications of computer vision. pp. 697--706 (2021)

\bibitem{zhong2020image}
Zhong, X., ShafieiBavani, E., Jimeno~Yepes, A.: Image-based table recognition:
  data, model, and evaluation. In: European conference on computer vision. pp.
  564--580. Springer (2020)

\bibitem{zhu2020deformable}
Zhu, X., Su, W., Lu, L., Li, B., Wang, X., Dai, J.: Deformable {DETR}:
  Deformable transformers for end-to-end object detection. In: ICLR (2021)

\end{thebibliography}

\clearpage

\section*{Appendix}

\appendix

\section{Licenses}

We plan to open source models and code under the MIT license.

\section{Accuracy vs.\ Page Table Cell Count}
\label{sec:appendix_accuracy_volume}

\begin{figure}[t]
\centering
\includegraphics[width=0.75\columnwidth]{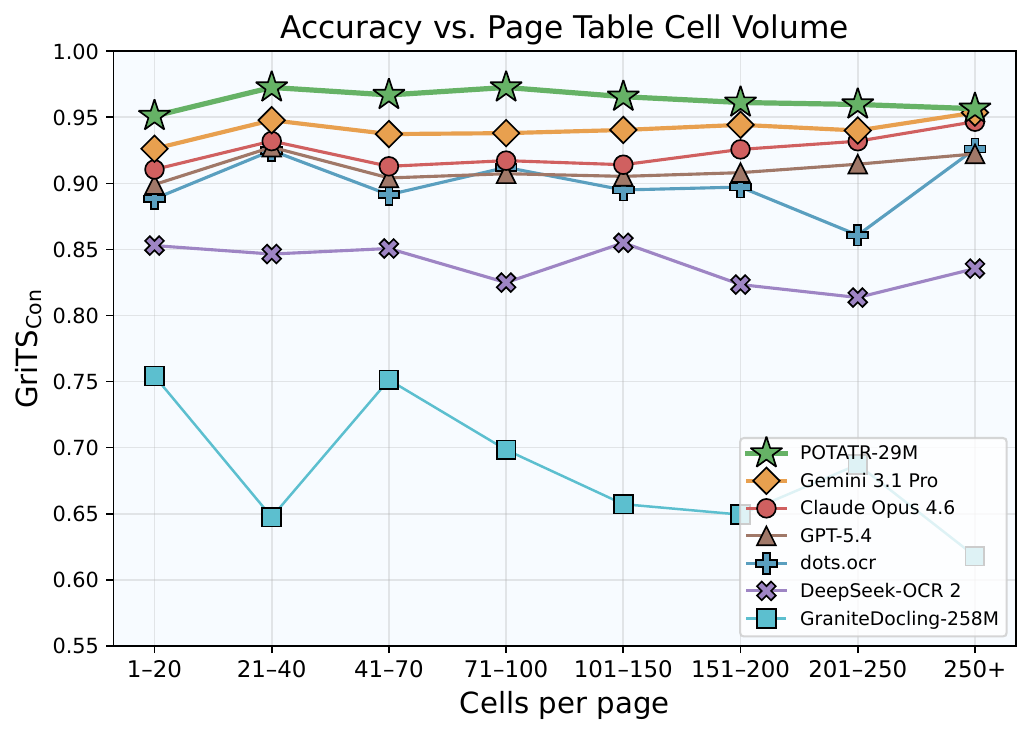}
\caption{\textbf{Accuracy vs.\ page table cell count.} $\textrm{GriTS}_\textrm{Con}$ for each model broken down by cells-per-page on PubTables-v2 Single Pages.}
\label{fig:accuracy_vs_volume}
\end{figure}

We break down $\textrm{GriTS}_\textrm{Con}$ by the number of table cells per page (\cref{fig:accuracy_vs_volume}).
Interestingly, no major trends in recognition performance emerge across models.
In fact, the frontier MLLMs show somewhat increased table extraction performance as the number of table cells grows.
This is despite the fact that the number of tokens generated, and thus the cost of output, significantly increases with page table cell count (\cref{fig:generation_cost}).
Furthermore, as the number of previous tokens in a model's context window grows with page table cell count, the amount of previous---potentially distributed---information on which the current token's prediction might need to be conditioned grows.
We speculate that this apparent paradox in performance could be because the page layout becomes more uniform (simpler) overall as table cell count on the page increases.
Thus any increase in the probability of error due to the increased token output is offset by the increased regularity in structure.
On the other hand, POTATR, which is not autoregressive, exhibits potentially more expected behavior, with performance peaking on the most typical tables in the data distribution.

\section{Multi-Page Table Extraction}
\label{sec:merging_results}

\begin{table}[tb]
\caption{\textbf{Multi-page table extraction.} Evaluation on PubTables-v2 Full Documents containing multi-page tables. Full-document models receive all pages at once, whereas page-by-page models process each page independently. The last two rows show the effect of applying cross-page merging \cite{smock2025pubtablesv2} to the page-by-page models' output.}
\centering
\begin{adjustbox}{width=\textwidth}
\begin{tabular}{l r r r r r r}
\toprule
\textbf{Model/Context} & $\textbf{GriTS}_{\textbf{Top}}$ & $\textbf{GriTS}_{\textbf{Con}}$ & $\textbf{Acc}_{\textbf{Top}}$ & $\textbf{Acc}_{\textbf{Con}}$ & $\textbf{TEDS}_{\textbf{S}}$ & $\textbf{TEDS}$ \\
\midrule
\multicolumn{7}{l}{\textbf{Full document}} \\
\quad Claude Opus 4.6 & 0.9116 & 0.9079 & \textbf{0.5799} & \textbf{0.2452} & 0.8914 & 0.8798 \\
\quad GPT-5.4 & 0.8865 & 0.8847 & 0.5038 & 0.1636 & 0.8563 & 0.8341 \\
\quad Gemini 3.1 Pro & \textbf{0.9365} & \textbf{0.9309} & 0.4854 & 0.2025 & \textbf{0.9048} & \textbf{0.8905} \\
\multicolumn{7}{l}{\textbf{Pages processed individually}} \\
\quad dots.ocr & 0.6054 & 0.5768 & 0.3958 & 0.1180 & 0.6179 & 0.5876 \\
\quad POTATR & 0.6974 & 0.6713 & 0.3289 & 0.2038 & 0.6354 & 0.6130 \\
\multicolumn{7}{l}{\textbf{Pages processed individually + cross-page merging}} \\
\quad dots.ocr + merging & 0.7772 & 0.7495 & 0.5059 & 0.1180 & 0.8022 & 0.7731 \\
\quad POTATR + merging & 0.8570 & 0.8348 & 0.4360 & 0.2238 & 0.8202 & 0.7998 \\
\bottomrule
\end{tabular}
\label{tab:experiment_multi_page_table_extraction}
\end{adjustbox}
\end{table}

In this experiment, we assess the gap in performance when applying POTATR, optimized for single-page TE, to the most challenging scenario for document-level TE---documents containing multi-page tables.

Initially we apply POTATR to these documents using page-by-page processing only.
This leads to an overall $\textrm{GriTS}_\textrm{Con}$ score of 0.671.
While POTATR can recognize single-page tables in these documents, any predicted table corresponds to at most only a fraction of any multi-page ground truth table, significantly lowering the achievable $\textrm{GriTS}_\textrm{Con}$ score.
For reference, we also compare to the best-performing small VLM, dots.ocr, which is also single-page only and achieves a $\textrm{GriTS}_\textrm{Con}$ score of 0.577.
Frontier MLLMs with full-document context, on the other hand, perform strongly: the best performing frontier model, Gemini 3.1 Pro, achieves $\textrm{GriTS}_\textrm{Con}$ of 0.931.

Next, we pair dots.ocr and POTATR with a ViT-B-16 table continuation classifier \cite{smock2025pubtablesv2}.
This model takes as input a pair of page images and predicts whether a table spans a page boundary.
When a single-page model predicts a table on both pages and the continuation model predicts a positive continuation, we merge the two table segments vertically.

POTATR with cross-page merging achieves $\textrm{GriTS}_\textrm{Con}$ of 0.835, a 49.7\% reduction in error over processing pages without merging (0.671).
Applying the same merging strategy to \texttt{dots.ocr} yields a 40.8\% reduction in error.
Notably, this improvement requires no retraining of the page-level model; it is achieved entirely through composition with a lightweight classifier (ViT-B-16, 86M parameters) and a simple vertical concatenation rule.
This addresses a substantial proportion of the gap between single-page models and frontier MLLMs on documents with multi-page tables---63\% for POTATR versus Gemini 3.1 Pro, and 77\% for POTATR versus GPT-5.4.

Cross-page table continuation prediction is near-optimal (0.995 recall), so the remaining gap is attributable to page-level extraction quality on multi-page table content (pages that POTATR never encounters during its current single-page training) and to within-page table part merging, which the current pipeline does not address.
Optimizing POTATR for these out-of-distribution pages, and extending the merging strategy to handle table parts split within a single page, are clear directions for future work.

\begin{credits}
\subsubsection{\discintname} The authors have no competing interests to declare that are relevant to the content of this article.
\end{credits}

\end{document}